\pdfoutput=1
\documentclass[11pt]{article}

\usepackage[]{ACL2023}

\usepackage{times}
\usepackage{latexsym}
\usepackage{booktabs}

\usepackage[T1]{fontenc}
\usepackage[utf8]{inputenc}

\usepackage{microtype}

\usepackage{inconsolata}

\usepackage{graphicx}
\usepackage{xcolor}
\usepackage{enumitem}
\usepackage{hyperref}
\usepackage{amsmath}

\usepackage{todonotes}

\newcommand{\strict}[1]{\textsc{Strict}}
\newcommand{\strictsmall}[1]{\textsc{Strict-small}}
\newcommand{\paper}[1]{\textsc{Paper}}
\newcommand{\vision}[1]{\textsc{Multimodal}}
\newcommand{\interact}[1]{\textsc{Interaction}}
\newcommand{\multilingual}[1]{\textsc{Multilingual}}

\title{BabyLM Turns 4 and Goes Multilingual:\\ Call for Papers for the 2026 BabyLM Workshop \\
 \vspace{0.2cm} \url{https://babylm.github.io/}}

\author{
BabyLM Team 
\AND
        Leshem Choshen \\ IBM Research, MIT \And  
        Ryan Cotterell \\ ETH Zürich \And  
        Mustafa Omer Gul \\ Cornell University \And
        Jaap Jumelet \\ University of Groningen   
        \AND
        Tal Linzen \\ NYU \And
        Aaron Mueller \\ Boston University \And 
        Suchir Salhan \\ University of Cambridge 
        \AND
        Raj Sanjay Shah \\ Georgia Tech \And  
        Alex Warstadt \\ UC San Diego \And
        Ethan Gotlieb Wilcox \\ Georgetown University
}
\begin{document}

\maketitle

\begin{abstract}
The goal of the BabyLM is to stimulate new research connections between cognitive modeling and language model pretraining.
We invite contributions in this vein to the BabyLM Workshop, which will also include the 4$^{\text{th}}$ iteration of the BabyLM Challenge.
As in previous years, the challenge features two ``standard'' tracks (\strict{} and \strictsmall{}), in which participants must train language models on under 100M or 10M words of data, respectively.
This year, we move beyond our previous English-only pretraining datasets with a new \multilingual{} track, focusing on English, Dutch, and Chinese.
For the workshop, we call for papers related to the overall theme of BabyLM, which includes training efficiency, small-scale training datasets, cognitive modeling, model evaluation, and architecture innovation.

\end{abstract}

\section{Introduction: BabyLM  %Suchir update
    }

The goals of BabyLM are to bring together multiple disciplines to answer an enduring question: how can a computational system learn language from limited inputs? Cognitive scientists investigate this question by trying to understand how \emph{humans} learn their native language during childhood. Computer scientists tackle this question by attempting to build efficient machine-learning systems to accomplish this task. 
% BabyLM brings these two communities together by asking how insights from cognitive science can be used to assemble more sample-efficient language models and how language modeling architectures can inspire research in cognitive science.
BabyLM aims to bridge these communities, encouraging the integration of insights from cognitive science into the design of more sample-efficient language models, while also using advances in language modeling architectures to generate new hypotheses and experimental paradigms for cognitive science.

BabyLM acts as both a dedicated challenge and a workshop, for the fourth year \citep{warstadt-etal-2023-findings,conll-2024-babylm,charpentier-etal-2025-findings}. We call for submissions of original research papers at the intersection of cognitive science and language modeling, and highlight this year's workshop theme: \textbf{\textit{Going beyond English}}. 
Motivated by the recent release of \textsc{BabyBabelLM} \citep{DBLP:journals/corr/abs-2510-10159}, we encourage workshop submissions to engage with topics related to multilingual language acquisition and typological diversity.

For the BabyLM challenge, we again challenge participants to train language models on human-sized amounts of data, up to 100 million words. %(See suggested topics in \S\ref{sec:topics}).
% \todo{say something about the theme}
This year's iteration will remain largely the same as previous editions, but with a few key differences, which we list below:
% \todo{write multilinguality section}
\begin{itemize}[leftmargin=0.5cm,rightmargin=0.5cm]
    \item We are debuting a new \multilingual{} track, in which participants are tasked to train models on a multilingual dataset taken from \textsc{BabyBabelLM} \citep{DBLP:journals/corr/abs-2510-10159}. More details are provided in \S\ref{sec:track_rules}. 
    \item We continue to offer the \strict{} and \strictsmall{} tracks, challenging participants to train on 100M and 10M words, respectively. The intermediate checkpointing requirement and the compute limitation that models may not conduct more than 10 epochs over their training data, both introduced in the past year, still remain. See \S\ref{sec:training-requirements} for details on how these requirements are instantiated for each track.
    \item In order to streamline the competition structure, we are removing the \vision{} and \interact{} tracks as dedicated tracks in the competition. These tracks are instead subsumed into the \strict{} and \strictsmall{} tracks. As a result, participants may also use paired image-text data or leverage feedback from a separate teacher model during training for \strict{} or \strictsmall{}. More details can be found in \S\ref{sec:track_rules}. 
\end{itemize}

\section{Key Dates} % Jaap

\emph{Tentative Timeline}: We will accept submissions through ACL Rolling Review (ARR) or directly through OpenReview. Paper submissions to the workshop can ignore competition entry deadlines. This is our anticipated timeline. Dates will be determined based on eventual ARR and conference deadlines.

\begin{itemize}[leftmargin=0.5cm,rightmargin=0.5cm,itemsep=0em]
    \item \textbf{Mid February:} CfP and training data released
    \item \textbf{Early April:} Evaluation pipeline and baselines released
    \item \textbf{May 25:} ARR submission deadline
    \item \textbf{Mid July:} Direct submissions deadline
    \item \textbf{Early August:} Direct submission reviews due, ARR commitment deadline
    \item \textbf{Mid August:} Decisions released
    \item \textbf{Early September:} Camera ready due
    \item \textbf{24--29 Oct:} Workshop @ EMNLP in Budapest (exact date TBA)  
\end{itemize}

\section{Non-competition Workshop Submissions} % suchir (leshem started)
\subsection{Topics} \label{sec:topics}
The BabyLM workshop encourages interdisciplinary submissions at the interface of language modeling, cognitive science, language acquisition, and/or evaluation. To this end, we will accept papers on a variety of topics, including but not limited to the following:
\begin{itemize}[noitemsep]
    \item Data-efficient architectures and training techniques.
    \item Data curation for efficient training.
    \item Cognitively and linguistically inspired language modeling and evaluation.
    \item Small models (and scale comparisons).
    \item Relevant aspects of multimodality.
    \item Interaction with or feedback from teacher models during training.
    \item Second language acquisition, bilingualism, or multilingualism.
\end{itemize}

\paragraph{Workshop Theme}
The workshop will have a theme this year: \textit{\textbf{Going beyond English}}.
Previous iterations of BabyLM have primarily focused on English, and with the introduction of the new \multilingual{} track we aim to inspire workshop submissions to investigate BabyLM-related topics with a focus on other languages.
We hope the \textsc{BabyBabelLM} dataset can serve as a starting point for this, band we also encourage submissions introducing new resources that will promote progress in data-efficient language modeling across diverse linguistic settings. 
In particular, we welcome work that expands coverage to underrepresented and low-resource languages, explores cross-lingual transfer and multilingual training strategies, and examines how typological diversity, morphology, and script variation shape learnability under BabyLM constraints. 
We are also interested in benchmarks, evaluation protocols, and ethical data practices that support inclusive and culturally grounded language technologies.

% \subsection{Formatting}
% The paper should follow the EMNLP format. This includes length and anonymity requirements upon submission. Reviewing will be double-blind.

\subsection{Paper submission} Submissions will be made through OpenReview. Submissions can be full archival papers (or non-archival upon request) and can be up to eight pages in length. Formatting requirements will follow standards for EMNLP 2026 workshops. This includes length and anonymity requirements upon submission. Reviewing will be double-blind. As in previous iterations, we do allow dual submission; however, we do not allow dual publication.

% LAST YEAR'S TEXT

%Submissions will be made through OpenReview. All submissions will be full archival papers and can be up to eight pages in length. Formatting requirements for submissions will be announced at a l
% ater date, once the presentation venue has been finalized. As before, we do allow dual-submission, however, we do not allow dual publication.

\subsection{Review \& Publication}
Papers submitted to the BabyLM workshop will be evaluated via a double-blind peer review process on merit and relevance. For competition participants, acceptance will be based primarily on technical soundness: we plan to reject only competition submissions that make incorrect or unjustified claims, have significant technical issues, do not reveal sufficient methodological details for replication, or demonstrate only minimal time investment. Feedback will largely be directed toward improving submissions.

% LAST YEAR'S TEXT:

%BabyLM will hold its own review process. The acceptance criteria are based on soundness: We plan only to reject submissions that make incorrect or unjustified claims. Other feedback will be directed at the improvement of submissions.

\section{Competition Details} % 

\subsection{Track Rules}
\label{sec:track_rules}
The second BabyLM Challenge includes three competition tracks: \textbf{\strict}, \textbf{\strictsmall}, and \textbf{\multilingual}. 
\textbf{Note that participation in one of the competition tracks is \underline{not} a prerequisite for submitting to the workshop!}

\paragraph{New track: \multilingual{}}
We are introducing a new \multilingual{} track this year, in which participants have to train models on a multilingual mixture of `cognitively plausible' training data.
The training data and evaluation pipeline for this track come from \textsc{BabyBabelLM} \citep{DBLP:journals/corr/abs-2510-10159}, which contains cognitively plausible training data for 45 different languages.

For the \multilingual{} track, we focus on three languages: English, Dutch, and Chinese.
Focusing on the full range of \textsc{BabyBabelLM} would require training data beyond the scope of the BabyLM Challenge, and we therefore decided to focus on a small set of languages with broad evaluation data available. 
However, we highly encourage people to focus on other languages for their workshop submission: the three languages of the challenge track have been selected to streamline the challenge itself.
We selected English to allow for a direct comparison to the \strict{} and \strictsmall{} tracks, and to study the impact that multilingual data has on model behavior.
We selected Dutch and Chinese as a way to investigate questions related to cross-lingual transfer in two ways: Dutch being a `typologically similar' language to English with the same script, and Chinese being `typologically distant' to both Dutch and English and with a different script.

For training, we introduce the following rule: participants are allowed to create a custom data mixture of the three languages that adds up to 100M tokens in total.
Word counts per language are adjusted by their \textit{Byte Premium} \cite{arnett-etal-2024-bit}, which adjusts for variation in orthographic and morphological structure by measuring the UTF-8 encoded size needed to express a fixed amount of content.\footnote{Byte Premiums are scaled with respect to English (1.0); Dutch has a premium of 1.0516, Chinese has a premium of 0.9894. What this means in practice is that 100MB of English data is equivalent to 105MB of Dutch and 98.9MB of Chinese (parallel) data.}
This way, it is possible to optimize the training data itself, enabling experimentation with optimal data domains and language ratios.

Evaluation will be done on a mixture of tasks assessing \textit{functional} and \textit{formal} linguistic competence \citep{mahowald2024dissociating}.
% Evaluation is partly done by fine-tuning and partly by zero-shot prompting, depending on the nature of the task.
Depending on the task, evaluation will be conducted either through fine-tuning or via zero-shot prompting. Prior evaluations have shown that in-context learning abilities are challenging to instill into LMs of BabyLM size, which remains an open question that we keep encouraging participants to explore.
Specifics on the evaluation tasks will be announced when the baseline models are released.

\paragraph{Continuing Tracks} % omer
The key rules for \textbf{\strict} and \textbf{\strictsmall} remain unchanged from past competitions \citep{choshen2024call, charpentier-etal-2025-findings}.
We quote:
\begin{quote}
    The \strict{} and \strictsmall{} tracks require that submissions be trained on a corpus of 100M words or less (in \strict{}) and 10M words or less (in \strictsmall{}). These tracks do \textit{not} require that participants use the official BabyLM corpus, although we will still provide an updated version of this dataset for participants interested in using it. Models in these tracks will be evaluated on language-only evaluation tasks.
    
    % In the \vision{} track, participants will train multi-modal image-text models. Participants can use whatever training procedure they wish, as long as models can provide (pseudo) log-likelihoods to strings of text, conditioned on an image. Participants are free to use whatever data they wish, as long as the dataset is within a 100M word budget. To facilitate easier participation in this track, we will re-release a suggested multimodal dataset that consists of 50\% text-only and 50\% paired image--text data. Submissions to this track will be evaluated on language-only tasks, as well as multi-modal tasks.
\end{quote}

This year, we have merged the \vision{} and \interact{} tracks into the \strict{} and \strictsmall{} tracks. This is for two primary reasons: Firstly, there were insufficient submissions to warrant standalone tracks.  Secondly, none of the submissions in the \interact{} track outperformed models submitted to the \strict{} track, and none of the submissions from \vision{} outperformed our baselines. We will still allow participants to use multimodal data or interactive training procedures if they choose, but such data or learning procedures must conform to the rules of the \strict{} or \strictsmall{} tracks. 
% As we have added a newer \multilingual{} track, we are therefore making \vision{} and \interact{} become part of \strict{} and \strictsmall{} to streamline the competition.

The main consequence of this change is that we now allow \strict{} and \strictsmall{} models to also be trained using interactions between multiple agents during training.
We will distinguish between a \textbf{submission model}, i.e., the participants' entry into the competition, and an \textbf{external model}, i.e., a secondary model used in the training pipeline of the submission model but not submitted to the competition.
From a cognitive perspective, the inclusion of an external model allows participants to simulate interactions between language learners and their caregivers or peers, as language acquisition is not passive and involves active language use from the learner~\citep{bruner1985child, clark2020conversational}. In practice, the external model could be used in the service of the submission model's learning in many ways. For example, this could be as simple as generating synthetic data for the submission model or providing feedback to the submission model's outputs, which could be rendered in text or be scalar rewards for use in reinforcement learning~\citep{martins2025once}. 

As before, we require external models to come from a predetermined list of models available on the BabyLM website. These external models themselves may be fine-tuned or trained without restriction, however. For the \strict{} track, the submission model must be exposed to no more than 100M word tokens (multiple exposures allowed, e.g., epochs); this word count includes text generated by external models \textit{and} pre-existing corpora. Additionally, the submission model may not generate more than 100M word tokens to be used as input for the external model. If the external model provides scalar rewards for use in RL, we also limit the number of rewards it can generate to 100M. For \strictsmall{}, each 100M word or reward limit becomes a 10M word or reward limit instead. Finally, we note that \textbf{distillation is not allowed:} The external model's tokenizer, weights, hidden states, or output distribution cannot be revealed to the submission model. If participants wish to perform knowledge distillation~\citep{hinton2015distilling} and use the external model's output distribution, the training word count of the external model would count towards the submission model's limit.

\subsection{Training Requirements}\label{sec:training-requirements} %raj
% \paragraph{Intermediate Checkpoints}

%\textbf{TODO} -- update this language to being a suggestion for epoch limitation; save a checkpoint after  every 1M words up to 10M, then 10M words up to 100M for a total of N checkpoints

% New this year, for deeper analysis, we will require submitting intermediate checkpoints. These checkpoints should be saved (i) every 1M words up to 10M words (ii) 10M words up to 100M words and (iii) every 100M words thereafter.

% \paragraph{Epoch Limitation}
As in last year's challenge, we have epoch limits and require the submission of intermediate checkpoints:
\begin{enumerate}
    \item Models submitted to the leaderboard can only be exposed to a fixed amount of input.
    \item Intermediate model checkpoints must be submitted as well to test for learning speed and model behavior dynamics.
\end{enumerate}
Below, we explain these choices in more detail.

\paragraph{Training Duration Limitations}
Similar to last year, we require the submission of results for models trained on a fixed amount of input (\emph{counting} repeated exposures), in particular \textbf{after at most 100M words for the \strictsmall{} track and after at most 1B words for all other tracks}.
In most cases, this will mean after 10 epochs on the standard BabyLM corpora. 
However, because what counts as an epoch may differ across submissions, we instead quantify training by the number of whitespace-separated input words. For the \multilingual{} track, this value should be computed while taking each language's Byte Premium into account~\citep{arnett-etal-2024-bit}.
While participants are welcome to train for longer and report this in their paper, we will only include the model checkpoints that follow these limitations in the leaderboard.
Note that it is also allowed to submit a model that is trained on less data: the 100M and 1B word limits are \textit{an upper bound} on data exposure. For the \strict{} and \strictsmall{} tracks, the number of words seen by the submission model is considered to be the sum of the number of input words and generated tokens.

%the number of words seen by the submission model is considered to be the number of input words from external sources, such as the BabyLM corpus or teacher generations, and excludes the student's own outputs. If a challenge submission incorporates interaction only after an initial pre-training stage (that is, if a student model is first trained up to the limit of 1B words and is then further finetuned via RL using only its own generated outputs), then checkpoints before and after the interaction process should be provided. In this case, the final model checkpoint following interactive training would be the leaderboard entry.

% Note that this is a minimal requirement: Results from additional checkpoints are also encouraged, in which case we recommend evaluating at exponentially growing intervals (e.g., 100K, 316K, 1M, ...).
% We will also allow entrants to submit only a single checkpoint at 100M words (or 10M words for \strictsmall), or a checkpoint at 100M (or 10M) words and a second checkpoint at less than 1B (or 100M) words.
% Results from models trained on more than 1B words can be submitted but will not count towards competition performance.

\paragraph{Intermediate Checkpoints} %omer
As in last year's challenge, we require the submission of intermediate model checkpoints to the HuggingFace Hub.
These checkpoints will be used for the updated evaluation pipeline (\S\ref{sec:eval}), to measure aspects related to learning efficiency and language acquisition.
The checkpoints we require will be at increasing intervals: every 1M words until 10M words are seen, every 10M words until 100M words are seen, and (for the tracks other than \strictsmall{}) every 100M words until 1B words are seen.
More precise details about the evaluation of these intermediate checkpoints will be announced with the release of the evaluation pipeline.

\paragraph{Motivation} 
Before last year, we provided no such requirements. 
One motivation for continuing this is that the training dynamics of LMs can be compared to the learning trajectories of children, which is valuable from a cognitive modeling perspective.
Furthermore, one of the conclusions of the 2024 BabyLM Challenge was that more compute is correlated with higher performance.
This runs counter to the goals of BabyLM in several ways:
First, one goal of BabyLM is developmentally plausible training, but children do not experience repeated exposure to their input.
While we allow that memories of inputs could have an impact on learning beyond the initial exposure, we judge hundreds of repeated exposures to every input to be developmentally implausible.
Second, another goal of BabyLM is to democratize pretraining research, but large numbers of training epochs require greater computational resources that are not available to all participants.
As a consequence, well-funded or well-equipped research groups have a significant advantage if no limitation is applied.
This advantage does not disappear with this restriction, as well-funded groups may be able to afford more hyperparameter searches and prototyping, but these efforts will at least lead to training recipes that can be reproduced in future cycles by less well-funded groups.

We have \emph{not} chosen to restrict the amount of compute.
While such a restriction might be ideal from the perspective of democratization, it is less clear (but by some estimates unlikely) that BabyLM submissions exceed the computation available to children \citep{sandberg2008whole}.
Furthermore, a requirement to compute FLOPs is more technically demanding than one to count the amount of data seen, and it could deter participation with limited additional advantages.

Finally, this restriction only applies to competition entries.
Workshop papers are not required to include models with 10 or fewer training epochs, though this is, of course, encouraged.

\subsection{Updated Dataset}\label{sec:dataset} % suchir 

\begin{table*}[t]
    \centering
    \resizebox{\linewidth}{!}{
    \begin{tabular}{llrrr}
    \toprule
    Dataset & Description & \# Words (multimodal data) & \# Words (strict data) & \# Images\\
    \midrule
    Localized Narratives \cite{LocalizedNarratives} & Image Caption & 27M & -- & 0.6M \\
    Conceptual Captions 3M \cite{CC3M} & Image Caption & 23M & -- & 2.3M \\
    CHILDES \citep{macwhinney2000childes} & Child-directed speech & 15M & 28.4M& -- \\
    British National Corpus (BNC), dialogue portion & Dialogue & 4M & 7.6M& -- \\
    Project Gutenberg (children's stories) \citep{gerlach-2018-gutenberg} & Written English & 13M & 25.5M& -- \\
    OpenSubtitles \citep{lison-tiedemann-2016-opensubtitles2016} & Movie subtitles & 10M & 21.8M& -- \\
    Simple English Wikipedia & Written Simple English & 7M & 15.3M& -- \\
    Switchboard Dialog Act Corpus \citep{Stolcke-etal:2000} & Dialogue & $<$1M & 248K& -- \\
    \midrule
    \emph{Total} & -- & 100M & 100M & 2.9M\\
    \bottomrule
    \end{tabular}}
    \caption{Datasets for the strict track of the BabyLM competition. We release a text-only and multimodal dataset; either is allowed, as are outside corpora, so long as the total word count of the training corpus is less than 100M. Datasets have been detoxified following the procedure of \citet{trhlik2026bias} (\S\ref{sec:dataset}).
    % \textsuperscript{1}\url{https://google.github.io/localized-narratives/}\ \ \ \textsuperscript{2}\url{https://ai.google.com/research/ConceptualCaptions/download}
    \looseness=-1}
    \label{tab:data}
\end{table*}

Recent work from last year's \textsc{Interaction} Track found that BabyLMs trained on the provided \textsc{Strict} corpus can produce toxic generations and hate speech, which was a proven challenge for sustainable multi-turn generations between a BabyLM and a Teacher model  \citep{salhan-etal-2025-teacher}. Subsequent analysis by \citet{trhlik2026bias} finds that the BabyLM corpus is more toxic and hateful than training corpora used in larger models such as BERT, with a comparatively higher presence of toxic and hateful sentences, even containing racist and sexualised training data, contrary to its child-aligned disposition. 

To address these issues, we release a modified detoxified training dataset based on the 2024 and 2025 training dataset \citep{choshen2024call} at this \href{https://osf.io/ad7qg/}{link}; see Table~\ref{tab:data} for dataset composition statistics.
This includes:
\begin{itemize}[noitemsep]
    \item 100M word \strict{} dataset
    \item 10M word \strictsmall{} dataset
    \item 100M word + image \vision{} dataset.
    
\end{itemize}

\noindent The dataset can be accessed on \href{https://huggingface.co/collections/BabyLM-community/babylm-2026}{this Huggingface collection}.

%The \textsc{Strict-Detox} and \textsc{Strict-Small-Detox} training corpora  ... \todo{Mention Suchir's detoxified corpus}

The datasets we provide serve as a starting point for training your models.
We do, however, also allow participants to `swap' (part of) the training data with other data sources of your choice, while adhering to the 10/100M word constraint.
This allows for experimentation of more optimal training data; datasets that have been used in previous BabyLM editions have been made publicly available at the following \href{https://docs.google.com/spreadsheets/d/1R4spgWHdSkYDZceaXHOdj0c-wMTSrn7ny7kC1lOf0ko/edit?usp=sharing}{link}.

\subsection{Evaluation}\label{sec:eval} %jaap

% PREVIOUS YEAR'S TEXT:

As in previous years, we will distribute an open-source evaluation pipeline, building on top of our repository from the 2025 challenge. Much of the evaluation will continue to be based on zero-shot probability comparisons of two text sequences. This year's new track, \multilingual{}, will follow a similar logic to the existing evaluation for \strict{}. Specifically, participants will be evaluating their trained models on a combination of zero-shot and finetuning-based tasks for English, Dutch, and Chinese. More details about the evaluation pipeline and the set of tasks will be released subsequently. %\todo{Do we need any mention of what we are removing or whether there will be hidden tasks?}
Similar to previous years, we will release a set of \textit{hidden tasks} shortly before the submission deadline, to test the robustness of models on previously unseen multilingual phenomena and domains, thereby assessing generalization beyond the released evaluation tasks.

\subsection{Baselines}
We will release a series of baseline models. 
Similar to the previous year, we will release baselines based on the winning submissions from the last year. For the \strict{} and \strictsmall{} tracks, this continues to be GPT-BERT~\citep{charpentier2024gptbertboth}. As we have merged the \interact{} track into the \strict{} and \strictsmall{} tracks, we will also be releasing our SimPO-based~\citep{meng2024simpo} preference optimization baseline from last year's \interact{} track as a baseline for the \strict{} and \strictsmall{} tracks. Finally, we will continue to offer GPT-2 Small~\citep{radford2019language} as a naive, purely autoregressive baseline.

For the \multilingual{} track, we will be adapting our GPT-BERT and GPT-2 Small baselines to the new task setting. This will be done in a naive manner. Specifically, we will construct a 100M word dataset featuring English, Dutch and Chinese by subsampling an equal number of words from the respective dataset of each language, while taking into account each language's Byte Premium~\citep{arnett-etal-2024-bit}. We will then perform BPE tokenization on the resulting dataset. Training for GPT-BERT and GPT-2 Small will then proceed in the same fashion as in their \strict{} track counterparts. We note again that this is intended to be a naive baseline. Participants can use different ratios of data between languages (or use custom datasets), experiment with tokenization and make contributions on architecture, training objectives, etc. as is the case with the \strict{} and \strictsmall{} tracks.

As before, these baselines are meant to encourage participants to innovate and improve beyond existing models and approaches. 

% PREVIOUS YEAR'S TEXT:

%We will release a series of baseline models. As opposed to last year's baselines, which were trained relatively naively, this year's baseline will be based on the winning submissions from last year. For the \strict{} and \strictsmall{} tracks, we will release the following baselines:  GPT2 (decoder-only; \citealp{radford2019language}), LTG-Bert (encoder-only; \citealp{LTGBert}), and Contextualizer (decoder-only; \citealp{Contextualizer}). 
%For the \vision{} track, we will release the GIT~\cite{wang2022git} and Flamingo~\cite{alayrac2022flamingo} baselines.
%These baselines are meant to encourage participants to innovate and improve beyond existing models.

\subsection{Competition Submission %[Aaron, Raj]
}

%As with last year, submissions will be made using the Dynabench platform, which is an online platform for dynamic data collection and model benchmarking.\footnote{\url{https://dynabench.org/}}

Competition paper submissions will be made through OpenReview. This will include links to models and predictions, as well as links to custom datasets if applicable.

Model predictions for competition submissions must also be uploaded to a Huggingface leaderboard before the deadline. While submissions to the leaderboard after the deadline are allowed and will be displayed in the leaderboard, we will not consider these entries as participating in the competition.

\vspace{0.2cm}
\noindent\textbf{What you need to submit:}
% \vspace{-0.3cm}
\begin{itemize}[leftmargin=0.5cm,rightmargin=0.5cm, itemsep=0cm]
    \item A Huggingface link where we can download the model and its intermediate checkpoints.
    \item Model predictions in a format compatible with the evaluation pipeline and Huggingface leaderboard. Our evaluation pipeline will have scripts for collating predictions for this purpose.
    \item A datasheet describing the composition of the custom dataset and containing a download link (if not using a BabyLM-provided corpus). For the \multilingual{} track, please upload your training corpus containing your custom language mixture.
    \item When using an external model in the \strict{} and \strictsmall{} tracks, the fine-tuning and distillation data for the external model (if any), and any data generated by the submission or external model.
\end{itemize}
Specific details on how and where to submit these things exactly will be shared on the BabyLM Slack and GitHub repository\footnote{\url{https://github.com/babylm}}.

\section{FAQs %[Suchi r]
}
\paragraph{Can I do cool idea X?}
If it is interesting, innovative, or may result in important findings, we want you to try it! If you think the rules are holding you back from submitting to the competition, please reach out to the organizers. If we find that the idea does not fit the competition's rules, the submission can still be an interesting workshop paper.

\paragraph{Why doesn't BabyLM do cool idea X?}
Maybe we haven't thought about it; please reach out.

\paragraph{Can papers be submitted to multiple tracks?} 
Yes. For example, a single paper can describe models that are submitted separately to the \strict{} and \multilingual{} tracks. 

\paragraph{Can I submit a paper about my work?}
Yes, we require that \emph{all} competition submissions be accompanied by a paper, which can be up to eight pages in length (though it does not need to be). Papers will be published in an archival format. All papers can describe experiments and analyses beyond the scope of the competition.

\paragraph{Can I submit additional evaluation metrics?}
Yes, you may submit additional evaluation metrics alongside a competition model in the \strict{}, \strictsmall{}, and \multilingual{} tracks. This type of contribution is especially encouraged for workshop submissions.

Moreover, we accept analysis and insightful findings on previous submissions or related topics and especially welcome evaluations that work well for small models but evaluate meaningful aspects. If you believe you know of an evaluation that we should use throughout the competition, please contact us.

\paragraph{What training regimes are permitted?}
Any training objective/regime is permitted as long as the data restrictions are followed. If you use ancillary models, for example, in the case of reranking or data augmentation, the training data for these models is counted towards your 100M word budget. This applies to all tracks; so, for example, while you can use the external model to produce POS tags, you cannot use an off-the-shelf POS tagger in your pipeline.

For evaluation purposes, we require that the model provide a function to score a sequence of words without the need for additional fine-tuning.

\paragraph{Are there any limits on hyperparameters?}
No. But please share at the end what you found so we can learn from your efforts.

\paragraph{Are there any limits on the number of epochs?}
As of last year's challenge, yes. Refer to the ``Training Duration Limitation'' paragraph of Section \ref{sec:training-requirements} for more details.

%Models in the \strict{} and \strictsmall{} can train for only 10 epochs, maximum. We do not impose this limit for models in the \vision{} or \interact{} tracks.
%No. We put no restrictions on the number of epochs, for several reasons: First, from an engineering perspective, training LMs with SGD tends to require multiple epochs at these scales to achieve peak performance. Second, from a cognitive perspective, humans have a memory of linguistic experience and can continue to access and learn from these memories. Third, we try not to make a stand on implementations to allow the most freedom for innovation. Our internal results suggest, however, that under regular circumstances, over-training on more than a couple epochs give minor gains at most. 

\paragraph{Can I use external tools?}
Yes, but if they are learned on language, their tokens are counted towards the 100M. That means one can train on the same text, both a tokenizer, a parser, an LM, etc., or on parts of the 100M, but the sum of all text seen by all training can not surpass the amount of text allowed. This raises the question of synthetic data, which is allowed under some restrictions. You may generate the 100M tokens in any legal way you like (yes, distilling or writing your own is fair, if you figure out what text facilitates learning, it is interesting regardless of how to gather such text), you may also train eventually on more than 100M words by augmentation, however, that only works in a closed system, i.e., the augmenters' training data counts toward the limit, so, for example, training two LMs on half of the words, and then having them generate more words and training a model on both the original data and the new one is legit (and it was not tested in the previous competition, so even the example itself is interesting).

\paragraph{I have different modalities that can help}
If it is not linguistic data, prove it, past years' submissions did not gain from non-linguistic grounding, but we encourage such scientific questions. If it is linguistic in nature (e.g., audio), then the words should still count towards the overall number of learned words.

\section{Organizing Committee}
(Alphabetical by last name) Leshem Choshen, Ryan Cotterell, Mustafa Omer Gul, Jaap Jumelet, Tal Linzen, Aaron Mueller, Suchir Salhan, Raj Sanjay Shah, Alex Warstadt, and Ethan Gotlieb Wilcox. Feel free to contact members of the organizing committee at: \texttt{leshem.choshen@mail.huji.ac.il}, \texttt{amueller@bu.edu}, \texttt{alexwarstadt@gmail.com}

\bibliography{custom}

@inproceedings{warstadt-etal-2023-findings,
    title = "Findings of the {B}aby{LM} Challenge: Sample-Efficient Pretraining on Developmentally Plausible Corpora",
    author = "Warstadt, Alex  and
      Mueller, Aaron  and
      Choshen, Leshem  and
      Wilcox, Ethan  and
      Zhuang, Chengxu  and
      Ciro, Juan  and
      Mosquera, Rafael  and
      Paranjabe, Bhargavi  and
      Williams, Adina  and
      Linzen, Tal  and
      Cotterell, Ryan",
    editor = "Warstadt, Alex  and
      Mueller, Aaron  and
      Choshen, Leshem  and
      Wilcox, Ethan  and
      Zhuang, Chengxu  and
      Ciro, Juan  and
      Mosquera, Rafael  and
      Paranjabe, Bhargavi  and
      Williams, Adina  and
      Linzen, Tal  and
      Cotterell, Ryan",
    booktitle = "Proceedings of the BabyLM Challenge at the 27th Conference on Computational Natural Language Learning",
    month = dec,
    year = "2023",
    address = "Singapore",
    publisher = "Association for Computational Linguistics",
    url = "https://aclanthology.org/2023.conll-babylm.1/",
    doi = "10.18653/v1/2023.conll-babylm.1",
    pages = "1--34"
}

@proceedings{conll-2024-babylm,
    title = "The 2nd BabyLM Challenge at the 28th Conference on Computational Natural Language Learning",
    editor = "Hu, Michael Y.  and
      Mueller, Aaron  and
      Ross, Candace  and
      Williams, Adina  and
      Linzen, Tal  and
      Zhuang, Chengxu  and
      Choshen, Leshem  and
      Cotterell, Ryan  and
      Warstadt, Alex  and
      Wilcox, Ethan Gotlieb",
    month = nov,
    year = "2024",
    address = "Miami, FL, USA",
    publisher = "Association for Computational Linguistics",
    url = "https://aclanthology.org/2024.conll-babylm.0/"
}

@inproceedings{charpentier-etal-2025-findings,
    title = "Findings of the Third {B}aby{LM} Challenge: Accelerating Language Modeling Research with Cognitively Plausible Data",
    author = "Charpentier, Lucas  and
      Choshen, Leshem  and
      Cotterell, Ryan  and
      Gul, Mustafa Omer  and
      Hu, Michael Y.  and
      Liu, Jing  and
      Jumelet, Jaap  and
      Linzen, Tal  and
      Mueller, Aaron  and
      Ross, Candance  and
      Shah, Raj Sanjay  and
      Warstadt, Alex  and
      Wilcox, Ethan Gotlieb  and
      Williams, Adina",
    editor = "Charpentier, Lucas  and
      Choshen, Leshem  and
      Cotterell, Ryan  and
      Gul, Mustafa Omer  and
      Hu, Michael Y.  and
      Liu, Jing  and
      Jumelet, Jaap  and
      Linzen, Tal  and
      Mueller, Aaron  and
      Ross, Candace  and
      Shah, Raj Sanjay  and
      Warstadt, Alex  and
      Wilcox, Ethan Gotlieb  and
      Williams, Adina",
    booktitle = "Proceedings of the First BabyLM Workshop",
    month = nov,
    year = "2025",
    address = "Suzhou, China",
    publisher = "Association for Computational Linguistics",
    url = "https://aclanthology.org/2025.babylm-main.28/",
    doi = "10.18653/v1/2025.babylm-main.28",
    pages = "399--420",
    ISBN = "TODO"
}

@inproceedings{arnett-etal-2024-bit,
    title = "A Bit of a Problem: Measurement Disparities in Dataset Sizes across Languages",
    author = "Arnett, Catherine  and
      Chang, Tyler A.  and
      Bergen, Benjamin",
    editor = "Melero, Maite  and
      Sakti, Sakriani  and
      Soria, Claudia",
    booktitle = "Proceedings of the 3rd Annual Meeting of the Special Interest Group on Under-resourced Languages @ LREC-COLING 2024",
    month = may,
    year = "2024",
    address = "Torino, Italia",
    publisher = "ELRA and ICCL",
    url = "https://aclanthology.org/2024.sigul-1.1/",
    pages = "1--9"
}

@article{mahowald2024dissociating,
  title = {Dissociating Language and Thought in Large Language Models},
  author = {Mahowald, Kyle and Ivanova, Anna A. and Blank, Idan A. and Kanwisher, Nancy and Tenenbaum, Joshua B. and Fedorenko, Evelina},
  year = {2024},
  month = mar,
  journal = {Trends in Cognitive Sciences},
  pages = {517--540},
  issn = {13646613},
  doi = {10.1016/j.tics.2024.01.011},
  url = {https://linkinghub.elsevier.com/retrieve/pii/S1364661324000275},
  urldate = {2024-03-22},
  langid = {english}
}

@article{DBLP:journals/corr/abs-2510-10159,
  author       = {Jaap Jumelet and
                  Abdellah Fourtassi and
                  Akari Haga and
                  Bastian Bunzeck and
                  Bhargav Shandilya and
                  Diana Galv{\'{a}}n{-}Sosa and
                  Faiz Ghifari Haznitrama and
                  Francesca Padovani and
                  Francois Meyer and
                  Hai Hu and
                  Julen Etxaniz and
                  Laurent Pr{\'{e}}vot and
                  Linyang He and
                  Mar{\'{\i}}a Grandury and
                  Mila Marcheva and
                  Negar Foroutan and
                  Nikitas Theodoropoulos and
                  Pouya Sadeghi and
                  Siyuan Song and
                  Suchir Salhan and
                  Susana Zhou and
                  Yurii Paniv and
                  Ziyin Zhang and
                  Arianna Bisazza and
                  Alex Warstadt and
                  Leshem Choshen},
  title        = {BabyBabelLM: {A} Multilingual Benchmark of Developmentally Plausible
                  Training Data},
  journal      = {CoRR},
  volume       = {abs/2510.10159},
  year         = {2025},
  url          = {https://doi.org/10.48550/arXiv.2510.10159},
  doi          = {10.48550/ARXIV.2510.10159},
  eprinttype    = {arXiv},
  eprint       = {2510.10159},
  bibsource    = {dblp computer science bibliography, https://dblp.org}
}

@article{choshen2024call,
  title={[Call for Papers] The 2nd BabyLM Challenge: Sample-efficient pretraining on a developmentally plausible corpus},
  author={Choshen, Leshem and Cotterell, Ryan and Hu, Michael Y and Linzen, Tal and Mueller, Aaron and Ross, Candace and Warstadt, Alex and Wilcox, Ethan and Williams, Adina and Zhuang, Chengxu},
  journal={arXiv preprint arXiv:2404.06214},
  year={2024}
}

@book{macwhinney2000childes,
  title={The CHILDES project: The database},
  author={MacWhinney, Brian},
  volume={2},
  year={2000},
  publisher={Psychology Press}
}

@article{gerlach-2018-gutenberg,
  doi = {10.48550/ARXIV.1812.08092},
  url = {https://arxiv.org/abs/1812.08092},
  author = {Gerlach, Martin and Font-Clos, Francesc},
  title = {A standardized {Project Gutenberg} corpus for statistical analysis of natural language and quantitative linguistics},
  journal = {Computing Research Repository},
  volume = {arXiv:1812.08092},
  year = {2018},
  copyright = {arXiv.org perpetual, non-exclusive license}
}

@inproceedings{lison-tiedemann-2016-opensubtitles2016,
    title = "{O}pen{S}ubtitles2016: Extracting Large Parallel Corpora from Movie and {TV} Subtitles",
    author = {Lison, Pierre  and
      Tiedemann, J{\"o}rg},
    booktitle = "Proceedings of the Tenth International Conference on Language Resources and Evaluation ({LREC}'16)",
    month = may,
    year = "2016",
    address = "Portoro{\v{z}}, Slovenia",
    publisher = "European Language Resources Association (ELRA)",
    url = "https://aclanthology.org/L16-1147",
    pages = "923--929",
}

@article{Stolcke-etal:2000,
    Author = {Stolcke, Andreas and Ries, Klaus and Coccaro, Noah and Shriberg, Elizabeth and Bates, Rebecca and Jurafsky, Daniel and Taylor, Paul and Martin, Rachel and Meteer, Marie and Van Ess-Dykema, Carol},
    Journal = {Computational Linguistics},
    Number = {3},
    Pages = {339--371},
    Title = {Dialogue Act Modeling for Automatic Tagging and Recognition of Conversational Speech},
    Volume = {26},
    Year = {2000}
}

@inproceedings{LocalizedNarratives,
  title={Connecting vision and language with localized narratives},
  author={Pont-Tuset, Jordi and Uijlings, Jasper and Changpinyo, Soravit and Soricut, Radu and Ferrari, Vittorio},
  booktitle={Computer Vision--ECCV 2020: 16th European Conference, Glasgow, UK, August 23--28, 2020, Proceedings, Part V 16},
  pages={647--664},
  year={2020},
  organization={Springer}
}

@inproceedings{CC3M,
  title={Conceptual captions: A cleaned, hypernymed, image alt-text dataset for automatic image captioning},
  author={Sharma, Piyush and Ding, Nan and Goodman, Sebastian and Soricut, Radu},
  booktitle={Proceedings of the 56th Annual Meeting of the Association for Computational Linguistics (Volume 1: Long Papers)},
  pages={2556--2565},
  year={2018}
}

@article{radford2019language,
  title={Language models are unsupervised multitask learners},
  author={Radford, Alec and Wu, Jeffrey and Child, Rewon and Luan, David and Amodei, Dario and Sutskever, Ilya and others},
  journal={OpenAI blog},
  volume={1},
  number={8},
  pages={9},
  year={2019}
}

@inproceedings{
    meng2024simpo,
    title={Sim{PO}: Simple Preference Optimization with a Reference-Free Reward},
    author={Yu Meng and Mengzhou Xia and Danqi Chen},
    booktitle={The Thirty-eighth Annual Conference on Neural Information Processing Systems},
    year={2024},
    url={https://openreview.net/forum?id=3Tzcot1LKb}
}

@misc{charpentier2024gptbertboth,
      title={GPT or BERT: why not both?}, 
      author={Lucas Georges Gabriel Charpentier and David Samuel},
      year={2024},
      eprint={2410.24159},
      archivePrefix={arXiv},
      primaryClass={cs.CL},
      url={https://arxiv.org/abs/2410.24159}, 
}

@article{sandberg2008whole,
  title={Whole brain emulation: A roadmap},
  author={Sandberg, Anders and Bostrom, Nick},
  year={2008},
  publisher={Future of Humanity Institute}
}

@inproceedings{salhan-etal-2025-teacher,
    title = "Teacher Demonstrations in a {B}aby{LM}{'}s Zone of Proximal Development for Contingent Multi-Turn Interaction",
    author = "Salhan, Suchir  and
      Gu, Hongyi  and
      Rooein, Donya  and
      Galvan-Sosa, Diana  and
      Gaudeau, Gabrielle  and
      Caines, Andrew  and
      Yuan, Zheng  and
      Buttery, Paula",
    editor = "Charpentier, Lucas  and
      Choshen, Leshem  and
      Cotterell, Ryan  and
      Gul, Mustafa Omer  and
      Hu, Michael Y.  and
      Liu, Jing  and
      Jumelet, Jaap  and
      Linzen, Tal  and
      Mueller, Aaron  and
      Ross, Candace  and
      Shah, Raj Sanjay  and
      Warstadt, Alex  and
      Wilcox, Ethan Gotlieb  and
      Williams, Adina",
    booktitle = "Proceedings of the First BabyLM Workshop",
    month = nov,
    year = "2025",
    address = "Suzhou, China",
    publisher = "Association for Computational Linguistics",
    url = "https://aclanthology.org/2025.babylm-main.25/",
    doi = "10.18653/v1/2025.babylm-main.25",
    pages = "323--355",
    ISBN = "TODO"
}

@article{trhlik2026bias,
  title={Bias Dynamics in BabyLMs: Towards a Compute-Efficient Sandbox for Democratising Pre-Training Debiasing},
  author={Trhlik, Filip and Caines, Andrew and Buttery, Paula},
  journal={arXiv preprint arXiv:2601.09421},
  year={2026}
}

@article{clark2020conversational,
  title={Conversational repair and the acquisition of language},
  author={Clark, Eve V},
  journal={Discourse Processes},
  volume={57},
  number={5-6},
  pages={441--459},
  year={2020},
  publisher={Taylor \& Francis}
}

@article{bruner1985child,
  title={Child's talk: Learning to use language},
  author={Bruner, Jerome},
  journal={Child Language Teaching and Therapy},
  volume={1},
  number={1},
  pages={111--114},
  year={1985},
  publisher={SAGE Publications Sage UK: London, England}
}

@inproceedings{martins2025once,
  title={Once Upon a Time: Interactive Learning for Storytelling with Small Language Models},
  author={Martins, Jonas Mayer and Bashir, Ali Hamza and Khalid, Muhammad Rehan and Beinborn, Lisa},
  booktitle={Proceedings of the First BabyLM Workshop},
  pages={454--468},
  year={2025}
}

@article{hinton2015distilling,
  title={Distilling the knowledge in a neural network},
  author={Hinton, Geoffrey and Vinyals, Oriol and Dean, Jeff},
  journal={arXiv preprint arXiv:1503.02531},
  year={2015}
}
\bibliographystyle{acl_natbib}

\end{document}